\documentclass[10pt,twocolumn,letterpaper]{article}
\usepackage[pagenumbers]{cvpr}

\usepackage[utf8]{inputenc}
\usepackage{times}
\usepackage{epsfig}
\usepackage{graphicx}
\usepackage{amsmath}
\usepackage{amssymb}
\usepackage[cjk]{kotex}
\usepackage{algpseudocode}
\usepackage{algorithm}
\usepackage[position=b]{subcaption}
\usepackage{multirow}
\usepackage{setspace}
\usepackage[pagebackref,breaklinks,colorlinks]{hyperref}
\usepackage[capitalize]{cleveref}
\crefname{section}{Sec.}{Secs.}
\Crefname{section}{Section}{Sections}
\Crefname{table}{Table}{Tables}
\crefname{table}{Tab.}{Tabs.}

\def\cvprPaperID{*****} 
\def\confName{CVPR}
\def\confYear{2022}

\newcommand{\bbf}[1]{\mathbf{#1}}
\newcommand{\bbb}[1]{\mathbb{#1}}
\newcommand{\mcal}[1]{\mathcal{#1}}
\newcommand{\tbf}[1]{\textbf{#1}}
\algnewcommand{\LineComment}[1]{\State \(\triangleright\) #1}

\begin{document}
\begin{sloppypar}

\title{Robust Pedestrian Attribute Recognition Using Group Sparsity for Occlusion Videos}

\author{Geonu Lee$^1$ \hspace{1.5cm} Kimin Yun$^2$ \hspace{1.5cm} Jungchan Cho$^{1}$ \\
$^1$College of Information Technology, Gachon University, South Korea\\
$^2$Electronics and Telecommunications Research Institute (ETRI), South Korea\\
{\tt\small \{lkw3139, thinkai\}@gachon.ac.kr \ \  kimin.yun@etri.re.kr}
}

\maketitle
\begin{abstract}
Occlusion processing is a key issue in pedestrian attribute recognition (PAR). Nevertheless, several existing video-based PAR methods have not yet considered occlusion handling in depth. In this paper, we formulate finding non-occluded frames as sparsity-based temporal attention of a crowded video. In this manner, a model is guided not to pay attention to the occluded frame. However, temporal sparsity cannot include a correlation between attributes when occlusion occurs. For example, “boots” and “shoe color” cannot be recognized when the foot is invisible. To solve the uncorrelated attention issue, we also propose a novel group sparsity-based temporal attention module. Group sparsity is applied across attention weights in correlated attributes. Thus, attention weights in a group are forced to pay attention to the same frames. Experimental results showed that the proposed method achieved a higher $F_1$-score than the state-of-the-art methods on two video-based PAR datasets.
\end{abstract}

\section{Introduction}
\let\thefootnote\relax\footnotetext{*Corresponding author: Jungchan Cho (thinkai@gachon.ac.kr)}
Pedestrian attribute recognition (PAR) is a task that predicts various attributes of pedestrians detected by surveillance cameras. It is a human-searchable semantic description and can be used in soft biometrics for visual surveillance~\cite{wang2022pedestrian}. There have been several studies on this subject~\cite{liu2017hydraplus, zhao2019recurrent, li2016human, han2019attribute, liu2018localization} because of the importance of its applications, such as finding missing persons and criminals. However, the occlusion problem is still under-handled. 

Because other objects and persons cause occlusions on a pedestrian, it is impossible to resolve it based on a single image. However, a video contains more information about a pedestrian than compared to an image, allowing a model to leverage information from multiple frames. 
Let us imagine that the lower body of a pedestrian is occluded at some frames, but the other frames have a visible lower-body appearance. 
In this case, we must use only the information from the frame with the lower body visible rather than the one in which the lower body is occluded. 
Recently, Chen \etal~\cite{chen2019temporal} proposed a video-based PAR method that calculates temporal attention probabilities to focus on frames that are important for attribute recognition. However, this method concentrates on incorrect frames if a pedestrian is occluded by other objects or other people. 
We argue that recent studies have not yet considered occlusion analysis in depth. In this paper, we propose a novel method for improving the PAR performance in occlusion cases. 

\begin{figure}[t]
\centering
\includegraphics[width=0.9
\linewidth]{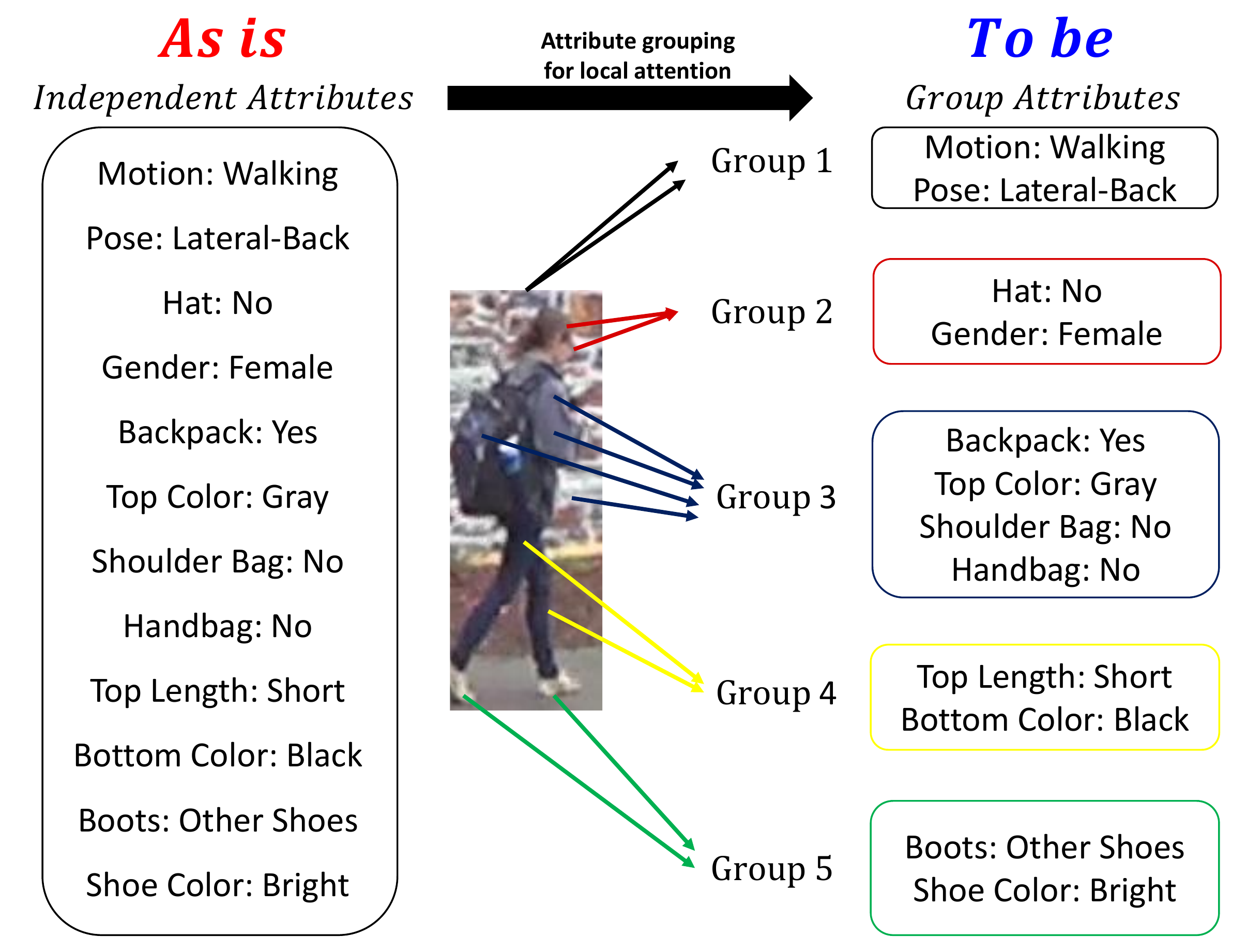}
\caption{Attribute grouping for local attention. Physically adjacent pedestrian attributes are grouped into one group. Group 1 is for attributes related to the entirety of a pedestrian. Groups 2, 3, 4, and 5 are for attributes related to the pedestrian's head, upper body, lower body, and feet, respectively. 
The network focuses on the semantic information of the pedestrian so that it can be robust against occluded pedestrians as obstacles.
}
\label{fig:grouping}
\end{figure}

As an intuitive idea, to avoid attending the frame with occlusion, we select the frame that can best estimate each attribute.
Therefore, one solution involves the use of sparsity regularization~\cite{friedman2017elements} of the temporal weights. In other words, sparse attention takes full advantage of the meaningful information in the other weighted frames. 
However, unfortunately, our experiment results showed that adding this simple sparsity constraint to the method in~\cite{chen2019temporal} cannot correctly handle occlusion. 
This is because the method proposed in~\cite{chen2019temporal} uses multiple independent branches for multi-attribute classification. On the other hand, pedestrian attributes are closely related to each other. In particular, semantically adjacent attributes have more significant relationships, as depicted in Figure~\ref{fig:grouping}. However, sparsity-constrained temporal attention cannot understand the relationships between the attributes either. 
Therefore, the relationship between attributes is key to finding meaningless frames, and we formulate it as a group sparsity-based temporal attention. 

\begin{figure*}[t]
\centering
\includegraphics[width=1.0\linewidth]{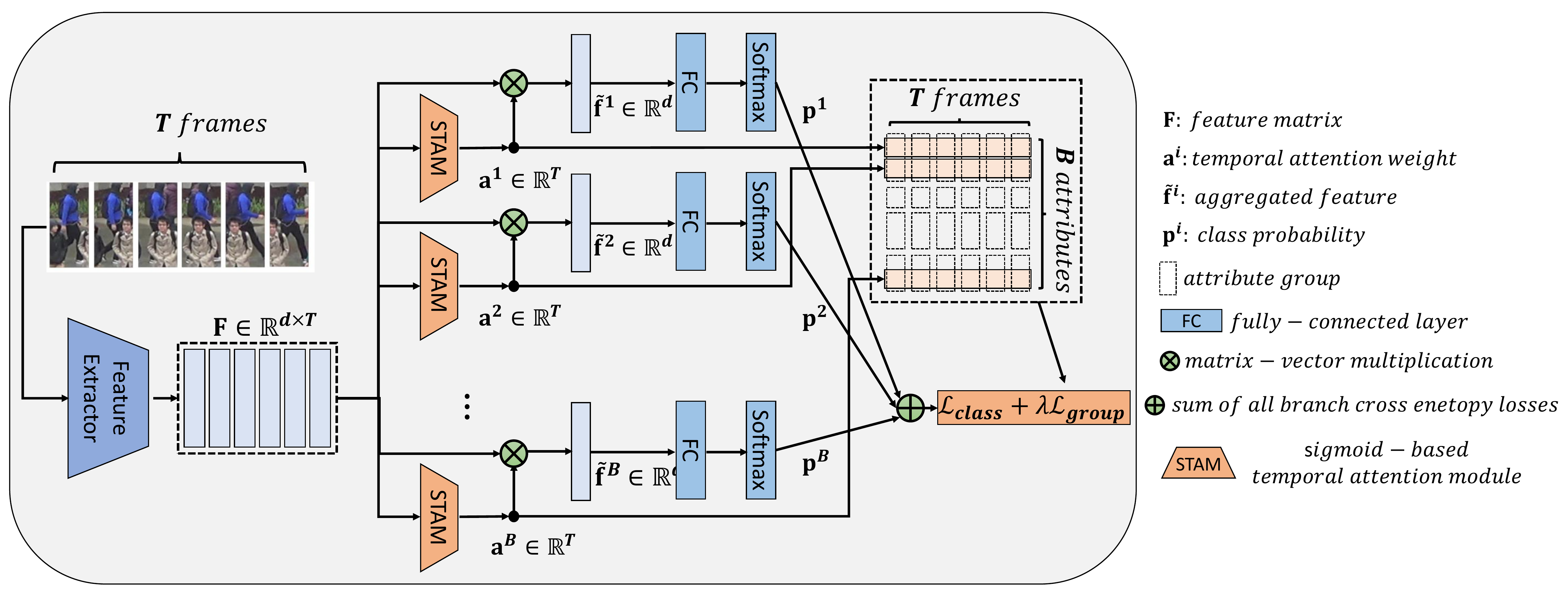}
\caption{Overview of the  network architecture of the proposed method. It consists of a feature extractor, sigmoid-based temporal attention modules, and attribute classifiers.
Because the attributes of the pedestrians are closely related to each other, the attention weights for semantically adjacent attributes have similar values to each other.
In other words, temporal frame attentions are not independent. 
To reflect this point, we formulate a group sparsity-based temporal attention module constraint.
}
\label{fig:overview}
\end{figure*}

Group sparsity~\cite{yuan2006model} is an advanced method compared to sparsity; it can gather the related attention of the attributes into a group. 
For instance, in Figure~\ref{fig:grouping}, information regarding the boots and shoe color is destroyed at the same time an obstacle occludes a pedestrian’s feet. 
In this case, group sparsity puts the boots and shoe color into one group. Then, their attention weights are simultaneously suppressed. Therefore, the group constraint achieves more robust results for occlusion situations than those of the sparsity method.  
Figure~\ref{fig:overview} represents an overview of the proposed method, which consists of a shared feature extractor, multiple attribute classification branches, and a group sparsity-based attention across multiple branches. Extensive experiments were conducted to demonstrate the robustness of the proposed method to occlusion. The proposed method achieved a higher $F_1$-score than that of the state-of-the-art methods on occlusion samples based on the DukeMTMC-VideoReID~\cite{chen2019temporal, wu2018exploit, ristani2016performance} and MARS~\cite{chen2019temporal, zheng2016mars} benchmark datasets.

Our main contributions are summarized as follows.
\begin{itemize}
    \item 
    The proposed temporal attention module is designed to reflect the temporal sparsity of useful frames in a crowded video. 
    Our model is guided to not pay attention to the occluded frame, but rather to the frame where relevant attributes are visible.
    \item
    When a pedestrian is occluded owing to obstacles, information on several related attributes is difficult to infer simultaneously. Therefore, we propose a novel group sparsity-based temporal attention module. This module allows a model to robustly pay attention to meaningful frames to recognize the group attributes of a pedestrian.
    \item
    Extensive experiments showed that the proposed method outperformed the state-of-the-art methods on two video-based PAR datasets, especially for occlusion samples.
\end{itemize}

The remainder of this paper is organized as follows. We first introduce some related work in Section~\ref{sec:related}. The proposed method is described in Section~\ref{sec:method}. Section~\ref{sec:experiments} presents the implementation details and experimental results. Finally, we conclude the paper in Section~\ref{sec:conclusion}.

\section{Related Works} \label{sec:related}
\subsection{Pedestrian Attribute Recognition}
Studies have been conducted on image-based PAR using various methods~\cite{liu2017hydraplus,zhao2019recurrent,li2019visual, tang2019improving}.
Liu~\etal~\cite{liu2017hydraplus} proposed the HydraPlus-Net network that utilizes multi-scale features.
Tang \etal~\cite{tang2019improving} proposed an attribute localization module (ALM) that learns specific regions for each attribute generated from multiple levels.
However, accurate attribute recognition for various environments such as occlusion situations is difficult to achieve with image-based PAR. 

A video has more information than that of an image; thus, the number of video-based studies has been increasing.
Chen \etal~\cite{chen2019temporal} proposed an attention module that indicates the extent to which the model pays attention to each frame for each attribute. They designed branches and classifiers for each attribute in the video.
Specker \etal~\cite{specker2020evaluation} used global features before temporal pooling to utilize the different pieces of information from various frames. However, existing video-based PAR methods have not yet considered occlusion problem in depth. In this paper, we focus on the occlusion handling of video-based PAR.

\subsection{Sparsity Loss}
The sparsity regularization is often used for selection problems~\cite{nguyen2018weakly,islam2021hybrid,rashid2020action,luo2021action}.
Nguyen \etal~\cite{nguyen2018weakly} proposed a sparse temporal pooling network for action localization in a video. In this method, the sparsity loss makes the model select the segments that are related to the target action.
Unlike the sparsity loss method that adjusts each value, the group sparsity loss method simultaneously controls the values associated with each other~\cite{cho2014robust, gao2015multi, yang2011tag, yang2013local, luo2013group, tan2017robust}.
We propose a method that adjusts the attention weights of pedestrian attributes at the same time by designing the group sparsity constraint.

\section{Proposed Method} \label{sec:method}

\begin{figure}[t] 
\centering
	\subfloat[]{
    \includegraphics[width=0.6\columnwidth]{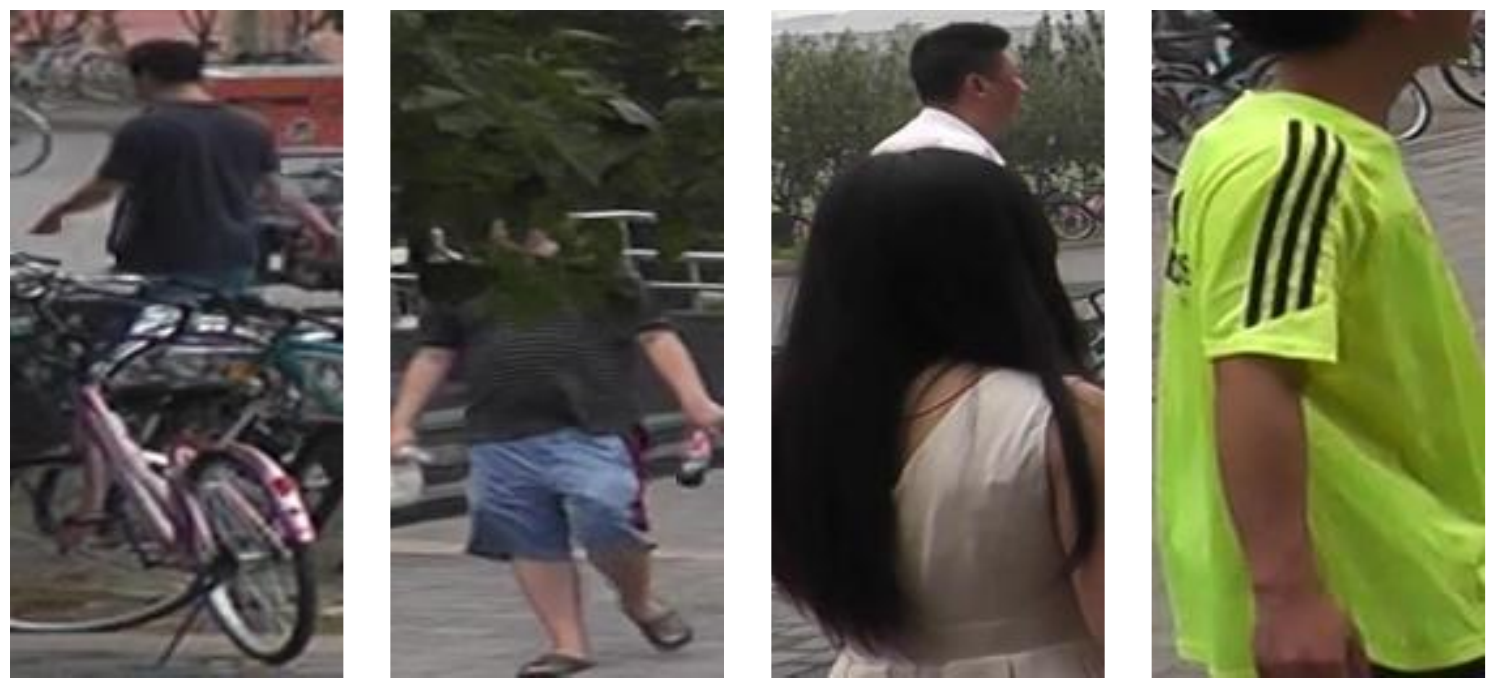}
    \label{fig:a:occ types}
    } 
    \subfloat[]{
    \includegraphics[width=0.375\columnwidth]{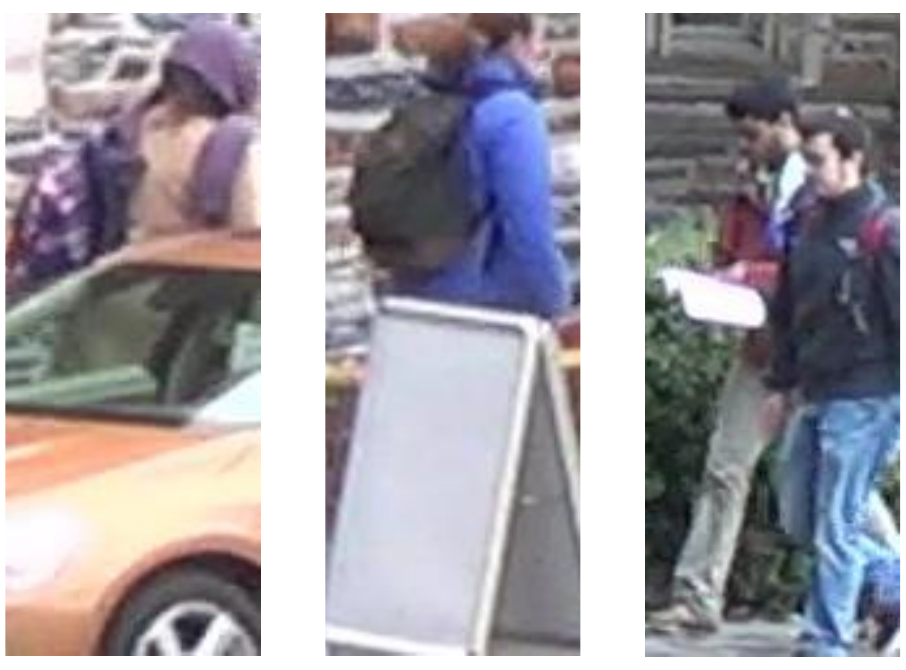}
    \label{fig:b:occ types}
    } 
\caption{(a) and (b) represent the occlusion types in MARS and DukeMTMC-VideoReID datasets, respectively. There are various occlusion types, such as a pedestrian’s lower body or head, other pedestrians, and tracking failure.}
\label{fig:occ types}
\end{figure} 

\subsection{Problem Formulation}
Figure~\ref{fig:occ types} shows examples of occluded pedestrian images from two video PAR datasets (DukeMTMC-VideoReID and MARS~\cite{chen2019temporal}). Typically, pedestrian images obtained from surveillance cameras in the real world are often obscured by crowded people, cars, and buildings. In addition, the instability of pedestrian tracking results in distorted pedestrian images. 
Therefore, it is important to robustly recognize the pedestrian attributes in occlusion situations; however, occluded pedestrian images make a single image-based PAR impossible. This study aims to achieve robust PAR using multiple frames, \ie, video-based PAR.

\subsection{Overview}
The proposed method consists of a feature extractor, attention modules, and attribute classifiers, and the inputs are a set of $T$ frames, as depicted in Figure~\ref{fig:overview}.
First, any feature extraction networks can be used. Here, we use the same feature extractor as that used in our baselines~\cite{chen2019temporal}, which consists of a ResNet~\cite{he2016deep} and two convolution modules to extract two types of feature according to their relevance to the identification (for more details, please see \cite{chen2019temporal}.). 
Second, a novel attention model is proposed to aggregate multiple features from $T$ frames in which attention weights are constrained by the temporal sparsity of frames and the group sparsity of pedestrian attributes. 
Finally, multi-branch classifiers are used for multi-labeled attribute classifications (\eg, hat, backpack, shoe type, and color). Notably, unlike the existing work~\cite{chen2019temporal}, which trains multiple attribute classifiers using independent classification loss only, the proposed method trains multiple classifiers using feature vectors constrained by a group sparsity-based temporal attention module. 
In the following sections, we will explain the novel group sparsity-based temporal attention module.

\subsection{Temporal Attention Module-Based Classification} \label{ssec:tam}
The body parts of a pedestrian are often occluded owing to obstacles and other pedestrians in real-world videos. Therefore, the information needed to recognize pedestrian attributes is different for each frame, even in the same video. 
For example, when there is a frame in which all parts of a pedestrian are visible and a frame in which an obstacle obscures the pedestrian’s lower body, the amount of information on the lower body provided by each frame is different.
We compute the temporal attention weights for $T$ frames to utilize the different pieces of information from each frame. 

Chen \etal~\cite{chen2019temporal} designed the temporal attention as a softmax-based probabilistic temporal attention module ($PTAM$) that calculates important probabilities for frames in the temporal direction. It consists of Conv-ReLU-Conv-ReLU-Softmax functions.
However, in the last ReLU-Softmax, ReLU~\cite{nair2010rectified} converts all the negative values to zero, and softmax normalizes the sum of the attention weights of the $T$ frame equal to 1. This makes it difficult to obtain attention weights that reflect sparsity constraints~\cite{nguyen2018weakly}. In other words, if the weight of a particular frame becomes 1, the weight of the rest of the frame becomes 0.
To solve this issue, we designed a sigmoid-based temporal attention module ($STAM$) configured with Conv-ReLU-Conv-Sigmoid. The sigmoid after Conv allows any frame to have a weight close to 0 or 1.

A temporal attention weight vector for the $i$-th attribute type, $\bbf{a}^{i} \in \bbb{R}^{T}$, can be obtained as
\begin{equation}
    \bbf{a}^{i} = STAM^{i}(\bbf{F}), 
\end{equation}
where $STAM^{i}(\cdot)$ is a sigmoid-based temporal attention module for the $i$-th attribute and $\bbf{F} = [\bbf{f}_1, \bbf{f}_2, \ldots, \bbf{f}_T] \in \bbb{R}^{d \times T}$ is a feature matrix that contains a set of $d$-dimensional feature vectors corresponding to $T$ frames, which is obtained from the same feature extractor as used in \cite{chen2019temporal}. 
Finally, an aggregated feature vector for the $i$-th attitude classification, $\bbf{\tilde{f}}^{i} \in \bbb{R}^{d}$, is obtained by multiplying the feature matrix $\bbf{F}$ and the attention weight vector $\bbf{a}^i$ as 
\begin{equation}
    \bbf{\tilde{f}}^{i} = \bbf{F}\bbf{a}^{i} 
                = \sum_{t=1}^{T} a^{i}_t\cdot\bbf{f}_t.
\label{eq:attention}
\end{equation}
Then, we pass $\bbf{\tilde{f}}^{i}$ to the $i$-th linear attribute classifier and return a prediction vector $\bbf{p}^{i}$ for each attribute.
\begin{equation}
    \bbf{p}^{i} = Softmax(\bbf{W}^{i}\bbf{\tilde{f}}^{i}),
\end{equation}
where $Softmax(\cdot)$ is a softmax function, $\bbf{W}^{i} \in \bbb{R}^{c \times d}$ is a weight matrix of a fully connected layer for the $i$-th attribute classification branch, and $c$ is the number of classes of the branch. The classification loss $\mcal{L}_{class}$ is the sum of the cross-entropy (CE)~\cite{goodfellow2016deep} of the attributes. 
\begin{equation}
    \mcal{L}_{class} = \sum_{i=1}^{B} \beta^i CE(\bbf{p}^{i}),
\end{equation}
where $B$ is the number of branches for each attribute in Figure~\ref{fig:overview}. 
$\beta^i$ is a balancing hyperparameter for the $i$-th attribute classification. 
It is set as a reciprocal of the number of classes in each attribute because each attribute classification has a different number of classes.

\subsection{Limitation of Sparsity Constraint on STAM} \label{ssec:whygroup}
The temporal attention weight $\bbf{a}^{i}$ in Equation~(\ref{eq:attention}) is an indicator that represents the importance of each frame. The sparsity constraint for the attention weight further emphasizes the effect and can be computed by the $\ell_1$-norm on $\bbf{a}^{i}$.
\begin{equation}
    \mcal{L}_{sparsity} = \sum_{i=1}^{B}{\lVert \bbf{a}^{i} \rVert}_{1},
\end{equation}
where $B$ is the number of branches of each attribute. Namely, the sparsity loss is the operation of the $\ell_1$ norm per branch of each attribute. From the formulation, the sparsity constraint is expected to have the effect of selecting frames that are not occluded from $T$ frames independently for each branch.

However, our experimental results presented in Section~\ref{sec:experiments} indicate that the sparsity constraint on the $STAM$ fails to make the correct frame importance, thereby degrading the PAR performance sometimes, as compared with the baselines. 
\\

\textit{Why does the sparsity constraint not improve the overall performance?}
\\

The sparsity constraint on $STAM$ is applied to the temporal attention weights by the $\ell_1$ norm, independently for each branch; thus, the attention weights of each branch depend only on the temporal information in each attribute. 
That is, the sparsity constraint does not help a model understand the relations between each attribute. 
However, pedestrian attributes are closely related to each other. As depicted in Figure~\ref{fig:occ types}, information about some attributes such as the type and color of the bottom and the type and color of shoe is damaged simultaneously if a pedestrian’s lower body or feet are occluded. 
Therefore, another constraint is needed to guide a model to understand the relationship between pedestrian attributes, which is important for achieving an algorithm that is robust to occlusion situations.
In the next section, we design the attribute relations as attribute groups and formulate the group constraints of attributes.

\subsection{Group Sparsity Constraint on STAM} \label{ssec:group}
Group sparsity extends and generalizes how to learn sparsity regularization, by which prior assumptions on the structure of the input variables can be incorporated~\cite{yuan2006model, obozinski2011group}. 
For the occluded pedestrian's attributes, the prior assumption is that pedestrian attributes can be partitioned into $K$ groups on the basis of their relevance,
\ie, $\mcal{G}^k$ where $k=1,2,\ldots, K$, as depicted in Figure~\ref{fig:grouping}. 
As a result, the attention weights in the same group at time $t$, $\{a^i_t|i \in \mcal{G}^k\}$, can be constrained by considering the group structure.

The method for grouping multiple attribute weights at time $t$ involves introducing a new vector at time $t$ using each attribute group, \ie, $\bbf{g}^k_t \in \bbb{R}^{|\mcal{G}^k|}$.
By computing the $\ell_2$ norm of a group vector $\bbf{g}^k_t$, we can define two sparsity constraints on attributes and time as 
\begin{equation}
    \mcal{L}_{group} = \sum_{t=1}^{T}\sum_{k=1}^{K} \gamma_k {{\lVert \bbf{g}^{k}_{t}} \rVert}_{2},
\end{equation}
where ${{\lVert \bbf{g}^{k}_{t}} \rVert}_{2}$ always has positive values and, thus the sum of these values is equal to the $\ell_1$ norm. $\gamma_k$ is a balancing hyperparameter for the $k$-th group in the sum of all the group sparsity loss functions. It is set as a reciprocal of the number of attributes in each group because each group has a different number of attributes.

The $\mcal{L}_{group}$ constraint on $STAM$ simultaneously increases or decreases the attention weights of specific groups in particular frames. 
It helps a model understand which frames are more important for each group and which groups in the same frame are recognizable. 
This constraint is consistent with the prior assumption that groups exist between attributes. 
In addition, it does not use explicit local patches in frames for specific attribute recognition. It uses implicit attention by attribute groups, enabling robust attribute recognition for pedestrian appearance distortions due to tracking failures.

Finally, the total loss function consists of $\mcal{L}_{class}$ and $\mcal{L}_{group}$ described above, as follows:
\begin{equation}
    \mcal{L}_{total} = \mcal{L}_{class} + \lambda \mcal{L}_{group}.
    \label{eq:loss}
\end{equation}
where $\lambda$ is a weight factor that combines the classification loss and the group sparsity loss.

\section{Experiments} \label{sec:experiments}

\subsection{Implementation Details} \label{ssec:details}
Tables~\ref{tb:group} show the attribute groups of the group sparsity for the experiments. We used the same feature extractor  as \cite{chen2019temporal}, which is pre-trained on the ImageNet dataset~\cite{deng2009imagenet}.
The initial learning rate was set to 3e-4 and multiplied by 0.3 at 100 epochs. The weight decay was set to 5e-4 for the Adam optimizer~\cite{kingma2014adam}. For the input, the width and height of the frame were resized to 112 and 224, respectively. The weight factor $\lambda$ in Equation~\ref{eq:loss} was set to 0.02. 
The batch size for training was set to 64. The model was trained for 200 epochs, and the best results were reported among the measurements every 20 epochs.  
The sequence length $T$ of the frames for training was set to six according to what was done in a previous work~\cite{chen2019temporal}. In the test phase, we divided the trajectory of a pedestrian into segments consisting of six frames. The divided segments were independently inferred, and the results were averaged for PAR. In other words, the performance was measured using one prediction per trajectory as done in~\cite{chen2019temporal}. 
We used a single NVIDIA Titan RTX GPU for both the training and the inference.
Regarding our experimental setting, if no additional explanation is given, we follow the process detailed in the baselines~\cite{chen2019temporal} for a fair comparison.

\subsection{Evaluation and Datasets}
We evaluated the proposed method using the average accuracy and $F_1$-score and compared it with four baselines: Chen \etal~\cite{chen2019temporal}, 3D-CNN~\cite{ji20123d}, CNN-RNN~\cite{mclaughlin2016recurrent}, and ALM~\cite{tang2019improving}. 
3D-CNN and CNN-RNN are video-based PAR methods compared in ~\cite{chen2019temporal}.
In the case of ALM~\cite{tang2019improving}, since it is an image-based PAR method, the image batch size was set to 96 and the learning rate was adjusted to 7.5e-5 according to ~\cite{goyal2017accurate}.
For a fair comparison, the random seed for the experiments was fixed deterministically and trained the baselines using the released codes.

For the extensive experiments, we used two video-based PAR datasets: DukeMTMC-VideoReID and MARS~\cite{chen2019temporal}, which were derived from the  re-identification datasets, DukeMTMC-VideoReID~\cite{wu2018exploit} and MARS~\cite{zheng2016mars}, respectively. Chen \etal~\cite{chen2019temporal} re-annotated them for the video-based PAR datasets.

\begin{table}[t]
\begin{center}
\caption{The attribute groups for DukeMTMC-VideoReID and MARS datasets.}
\label{tb:group}
\resizebox{\columnwidth}{!}{
\begin{tabular}{c|c|c}
\hline
\multirow{2}{*}{Group} & \multirow{2}{*}{\shortstack{DukeMTMC-VideoREID}} & \multirow{2}{*}{\shortstack{MARS}}\\
&&\\
\hline\hline
Whole & motion, pose & motion, pose\\
\hline
Head & hat, gender & age, hat, hair, gender \\
\hline
\multirow{2}{*}{Upper Body} & backpack, top color, shoulder bag, & backpack, top color, shoulder bag,\\
& handbag & handbag, top length\\
\hline
\multirow{2}{*}{Lower Body} & \multirow{2}{*}{top length, bottom color} & bottom length, bottom color,\\
&&type of bottom\\
\hline
Foot & boots, shoe color & -\\
\hline
\end{tabular}}
\end{center}
\end{table}

\subsubsection{DukeMTMC-VideoReID Dataset}
The DukeMTMC-VideoReID dataset contains 12 types of pedestrian attribute annotations.
The eight attributes are binary types: backpack, shoulder bag, handbag, boots, gender, hat, shoe color, and top length. 
The other four attributes are multi-class types: motion, pose, bottom color, and top color.
The attributes were annotated per trajectory, and the total number of trajectories was 4832. 
We excluded four trajectories with fewer frames than the segment length $T$, and the remaining 4828 trajectories were used in the experiments. 
For the training, 2195 trajectories were used, 413 of which contained occlusions, as shown in Figure~\ref{fig:b:occ types}.
For the test, 2633 trajectories were used, 449 of which contained occlusions.
The average length of the trajectories was approximately 169 frames.

\begin{table}[t]
\begin{center}
\caption{Comparisons of the results for the occlusion samples of the DukeMTMC-VideoReID and MARS datasets.}
\label{tb:occlusion}
\resizebox{\columnwidth}{!}{
\begin{tabular}{c|c|c|c}
\hline
\multirow{2}{*}{Dataset} & \multirow{2}{*}{Method} & Average & Average\\
&& Accuracy (\%) & $F_1$-score (\%)\\
\hline\hline
\multirow{5}{*}{\shortstack{DukeMTMC\\ -VideoReID}}  & Chen \etal~\cite{chen2019temporal} & 88.33 & 69.03 \\
                      & 3DCNN \cite{ji20123d} & 84.41 & 61.38 \\  
                      & CNN-RNN \cite{mclaughlin2016recurrent} & 87.94 & 68.12 \\
                      & ALM \cite{tang2019improving} & 86.99 & 65.87 \\
                      & Ours & \tbf{88.36} & \tbf{70.21} \\
                
\hline
\multirow{5}{*}{MARS}   & Chen \etal~\cite{chen2019temporal} & 66.39 & 55.67 \\
                        & 3DCNN \cite{ji20123d} & 60.83 & 46.16 \\ 
                        & CNN-RNN \cite{mclaughlin2016recurrent} & 65.83 & 53.79 \\
                        & ALM \cite{tang2019improving} & 67.50 & 55.73 \\
                        & Ours & \tbf{71.94} & \tbf{61.88} \\
\hline
\end{tabular}}
\end{center}
\end{table}

\subsubsection{MARS Dataset}

The MARS dataset contains 14 types of pedestrian attribute annotations.
The ten attributes are binary types: shoulder bag, gender, hair, bottom type, bottom length, top length, backpack, age, hat, and handbag.
The other four attributes are multi-class types: motion, pose, top color, and bottom color.
The attributes were also annotated per trajectory, and the total number of trajectories was 16,360. 
We also excluded five trajectories with fewer frames than the segment length $T$, and the remaining trajectories were 16,355. 
For the training, 8297 trajectories were used, 35 of which contained occlusions, as shown in Figure~\ref{fig:a:occ types}.
For the test, 8058 trajectories were used, 30 of which contained occlusions.
The average length of the trajectories was about 60 frames.

\subsection{Evaluation on the DukeMTMC-VideoReID and MARS Datasets}

To evaluate the robustness of the proposed method in occlusion situations, we compared its performance using only the occlusion samples with those of the baselines.
Table~\ref{tb:occlusion} presents the results on the DukeMTMC-VideoReID and MARS datasets.
To ensure accurate evaluations, we excluded the hat and handbag attributes of the MARS dataset because the ground truth of both attributes for all occlusion samples was the same, \ie, “no.”
As shown in Table~\ref{tb:occlusion}, the proposed method outperformed the baselines in all cases and achieved average accuracies of 88.36\% and 71.94\%, and average $F_1$-scores of 70.21\% and 61.88\% on the occlusion samples of the DukeMTMC-VideoReID and MARS datasets, respectively.

\begin{table}[t]
\begin{center}
\caption{Comparisons of the results for the total samples of the DukeMTMC-VideoReID and MARS datasets.}
\label{tb:total}
\resizebox{\columnwidth}{!}{
\begin{tabular}{c|c|c|c}
\hline
\multirow{2}{*}{Dataset} & \multirow{2}{*}{Method} & Average & Average\\
&& Accuracy (\%) & $F_1$-score (\%)\\
\hline\hline
\multirow{5}{*}{\shortstack{DukeMTMC\\ -VideoReID}}  & Chen \etal~\cite{chen2019temporal} & \tbf{89.12} & 71.58 \\
                      & 3DCNN \cite{ji20123d} & 85.38 & 64.66 \\  
                      & CNN-RNN \cite{mclaughlin2016recurrent} & 88.80 & 71.73 \\
                      & ALM \cite{tang2019improving} & 88.13 & 69.66 \\
                      & Ours & 88.98 & \tbf{72.30} \\

\hline
\multirow{5}{*}{MARS}   & Chen \etal~\cite{chen2019temporal} & 86.42 & 69.92 \\
                        & 3DCNN \cite{ji20123d} & 81.96 & 60.39 \\ 
                        & CNN-RNN \cite{mclaughlin2016recurrent} & 86.49 & 69.89 \\
                        & ALM \cite{tang2019improving} & 86.56 & 68.89 \\
                        & Ours & \tbf{86.75} & \tbf{70.42} \\
\hline
\end{tabular}}
\end{center}
\end{table}

Table~\ref{tb:total} shows the performances of the methods on the total samples of the DukeMTMC-VideoReID and MARS datasets, where the proposed method outperformed the baselines. Only in one case in the DukeMTMC-VideoReID dataset did the Chen \etal~\cite{chen2019temporal} method show slightly better average accuracy. However, because the measure of average accuracy did not consider data imbalance, the difference was negligible. 
In addition, Table~\ref{tb:total} does not correctly show the performances on the occlusion samples because the percentage of such samples among the total samples was very low.

\subsection{Ablation Study} \label{ssec:ablation}
\subsubsection{Effects of the Weight Factor $\lambda$}\label{sssec:lambda}
We compared the experimental results according to the weight factor $\lambda$ in Equation~\ref{eq:loss}. The weight factor $\lambda$ is a parameter that adjusts the sparsity. 
As shown in Table~\ref{tb:group sparsity occ f1}, the proposed method shows higher $F_1$-scores than those of the baseline methods, regardless of $\lambda$ values and the best results were obtained with $\lambda=0.02$.

\begin{figure*}[]
\centering
\includegraphics[width=1.0\linewidth, keepaspectratio=true]{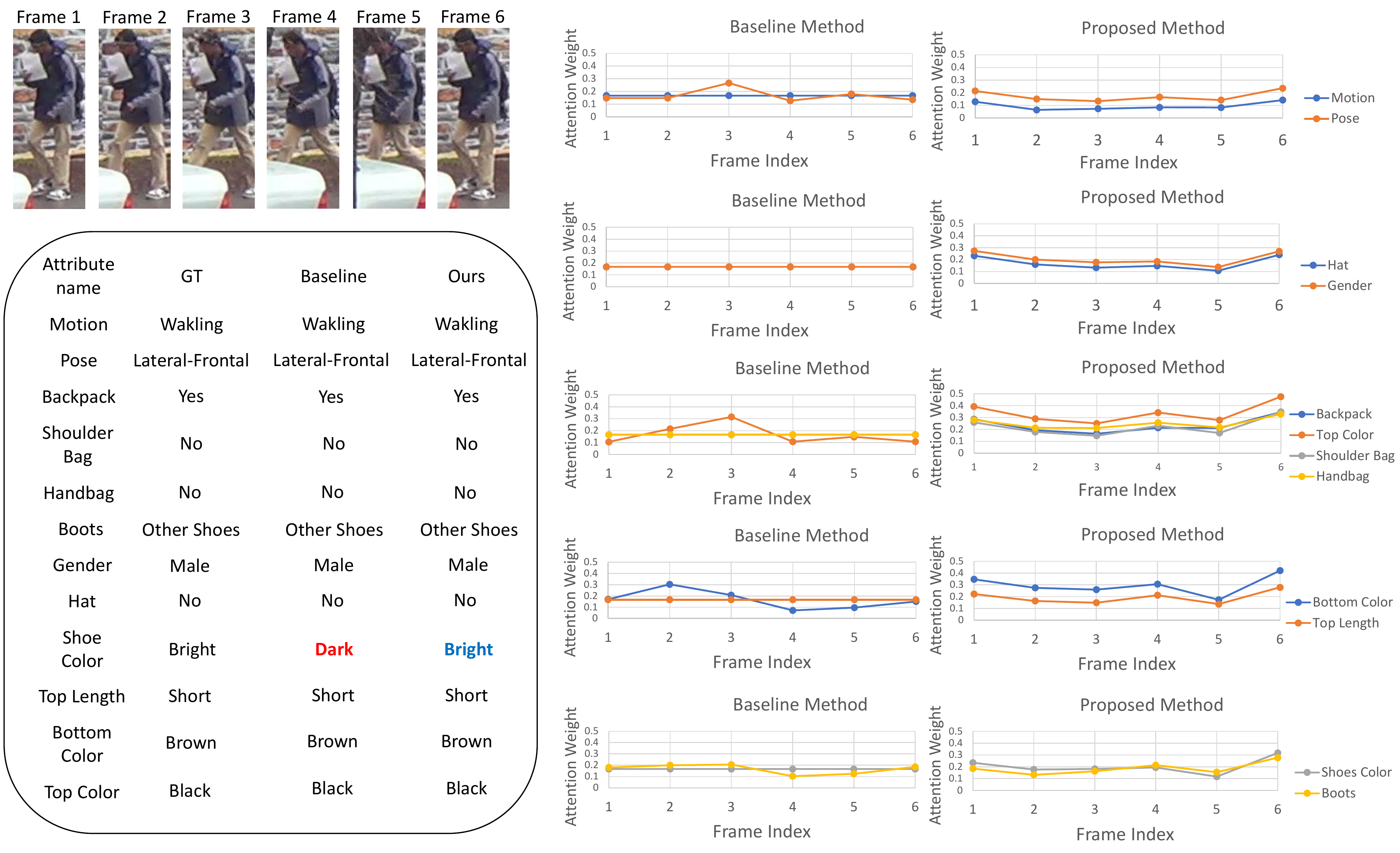}
\caption{
Qualitative results for the DukeMTMC-VideoReID dataset. It shows the attention weights of the group attributes and the PAR results. For the groups related to the lower body, the proposed method has low attention weights in the occluded frames. However, the attention weights of the baseline method (Chen~\etal~\cite{chen2019temporal}) are almost the same in all the frames.}
\label{fig:duke attention}
\end{figure*}

\begin{table}[t]
\begin{center}
\caption{Analysis of the group sparsity loss for the occlusion samples of the DukeMTMC-VideoReID and MARS datasets.}
\label{tb:group sparsity occ f1}
\resizebox{\columnwidth}{!}{
\begin{tabular}{c|c|c|c}
\hline
\multirow{2}{*}{Dataset} & \multirow{2}{*}{Method} & Average & Average\\
&& Accuracy (\%) & $F_1$-score (\%)\\
\hline\hline
\multirow{4}{*}{\shortstack{DukeMTMC\\ -VideoReID}} & Chen \etal~\cite{chen2019temporal} & 88.33 & 69.03 \\
                      & $\lambda$ = 0.005 & \tbf{88.38} & 69.85 \\  
                      & $\lambda$ = 0.03  & 88.16 & 69.62 \\
                      & $\lambda$ = 0.02 & 88.36 & \tbf{70.21} \\
\hline
\multirow{4}{*}{MARS}  & Chen \etal~\cite{chen2019temporal} & 66.39 & 55.67 \\
                      & $\lambda$ = 0.005 & 68.06 & 55.07 \\  
                      & $\lambda$ = 0.03  & 70.00 & 58.89 \\
                      & $\lambda$ = 0.02 & \tbf{71.94} & \tbf{61.88} \\
\hline
\end{tabular}}
\end{center}
\end{table}

\begin{table}[t]
\begin{center}
\caption{Comparisons between the sparsity-based and the group sparsity-based (ours) constraints for the occlusion samples of the DukeMTMC-VideoReID and MARS datasets.}
\label{tb:sparsity vs group occ f1}
\resizebox{1.0\columnwidth}{!}{
\begin{tabular}{c|c|c|c|c|c}
\hline
\multirow{2}{*}{Dataset} & \multirow{2}{*}{Method} & \multirow{2}{*}{PTAM} & \multirow{2}{*}{STAM} & Average & Average \\
 & & & & Accuracy (\%) & $F_1$-score (\%)\\
\hline\hline
\multirow{6}{*}{\shortstack{DukeMTMC\\ -VideoReID}}  & Chen \etal~\cite{chen2019temporal} & \checkmark & - & 88.33 & 69.03 \\
                      & Sparsity  & \checkmark & - & 87.99 & 69.05 \\
                      & Group sparsity  & \checkmark & - & 88.23 & \tbf{70.24} \\
                      & Chen \etal~\cite{chen2019temporal} & - & \checkmark & 87.94 & 69.26 \\
                      & Sparsity  & - & \checkmark & 87.68 & 67.52 \\  
                      & Group sparsity  & - & \checkmark & \tbf{88.36} & 70.21 \\
\hline
\multirow{6}{*}{MARS}  & Chen \etal~\cite{chen2019temporal} & \checkmark & - & 66.39 & 55.67 \\
                      & Sparsity  & \checkmark & - & 70.00 & 57.76 \\
                      & Group sparsity  & \checkmark & - & \tbf{71.94} & 61.70 \\ 
                      & Chen \etal~\cite{chen2019temporal} & - & \checkmark & 66.94 & 55.92 \\                   
                      & Sparsity  & - & \checkmark & 69.17 & 57.80 \\  
                      & Group sparsity  & - & \checkmark & \tbf{71.94} & \tbf{61.88} \\
\hline
\end{tabular}}
\end{center}
\end{table}

\subsubsection{Comparisons Between PTAM and STAM} \label{sssec:stam}
Table~\ref{tb:sparsity vs group occ f1} shows that the sparsity has the worst performance in terms of both accuracy and $F_1$-scores. As explained in Section~\ref{ssec:whygroup}, the sparsity constraint cannot help a model understand the relationship between attributes. 
However, the proposed method using the group sparsity-constrained STAM, which understands the relationship between each attribute, showed the best performance compared to the other methods.

\subsection{Qualitative Results}
We visualized the temporal attention weight vector with various segment frames to analyze the proposed method’s robustness to occlusion situations.
Figure~\ref{fig:duke attention} presents the temporal attention vectors and the PAR results of the method presented by Chen~\etal~\cite{chen2019temporal} and that of our method for all the groups of the DukeMTMC-VideoReID dataset. 
The values of the baseline method have similar values in all the frames.
In contrast, the values of the proposed method have different values in each frame.
Moreover, the values of the occlusion frames are lower than those of the general frames.
The attention weights of the bottom and top length attributes are simultaneously controlled because they belong to the same group. 
For the same reason, the attention weights of the shoe color and boot attributes are also simultaneously adjusted.
As a result, the baseline method predicted the shoe color attribute different from the ground truth. However, the proposed method accurately predicted all attributes.

\section{Conclusion} \label{sec:conclusion}
This paper proposed a novel group sparsity-constrained temporal attention module to robustly recognize pedestrian attributes in occlusion situations.
The proposed method was formulated as a group sparsity to consider the relationship between pedestrian attributes, which improves the temporal attention. The results of extensive experiments demonstrated that the proposed method consistently outperformed all the baselines.

\section*{Acknowledgments}
This work was supported in part by Institute of Information \& Communications Technology Planning \& Evaluation~(IITP) grant funded by the Korea government~(MSIT)~(No.2014-3-00123, Development of High Performance Visual BigData Discovery Platform for Large-Scale Realtime Data Analysis) and in part by the Gachon University research fund of 2020(GCU-202008450006).

{\small
\bibliographystyle{ieee_fullname}

}

\end{sloppypar}
\end{document}